# Curriculum Learning In Job Shop Scheduling Using Reinforcement Learning

Constantin Waubert de Puiseau[1], Hasan Tercan[1], Tobias Meisen[1]

[1]Chair for Technologies and Management of Digital Transformation, Wuppertal, Germany

**Abstract**

Solving job shop scheduling problems (JSSPs) with a fixed strategy, such as a priority dispatching rule, may yield satisfactory results for several problem instances but, nevertheless, insufficient results for others. From this single-strategy perspective finding a near optimal solution to a specific JSSP varies in difficulty even if the machine setup remains the same. A recent intensively researched and promising method to deal with difficulty variability is Deep Reinforcement Learning (DRL), which dynamically adjusts an agent's planning strategy in response to difficult instances not only during training, but also when applied to new situations. In this paper, we further improve DLR as an underlying method by actively incorporating the variability of difficulty within the same problem size into the design of the learning process. We base our approach on a state-of-the-art methodology that solves JSSP by means of DRL and graph neural network embeddings. Our work supplements the training routine of the agent by a curriculum learning strategy that ranks the problem instances shown during training by a new metric of problem instance difficulty. Our results show that certain curricula lead to significantly better performances of the DRL solutions. Agents trained on these curricula beat the top performance of those trained on randomly distributed training data, reaching 3.2% shorter average makespans.

## 1. Introduction

Inspired by the way humans learn, deep reinforcement learning (DRL) is a machine learning paradigm in which a system, or agent, autonomously learns from gathered experience. Most famously, DRL has been successfully applied to board and video games [1,2] with superhuman performance. In recent years, DRL has also shown promising results in industrial use-cases and combinatorial optimization problems such as the job shop scheduling problem (JSSP) [3–5].

Scheduling problems deal with the allocation of resources to jobs over time to optimize criteria such as total time spent to process all jobs, called makespan [6]. The JSSP in particular is a problem formulation, where each job must visit each machine in a factory in a fixed order, and is considered NP-hard to solve optimally. In practice, scheduling problems are often solved using priority dispatching rules (PDRs) consisting of simple rules for determining the priority of jobs over a scheduling sequence [6]. The main promise of DRL for the scheduling problems compared to alternative solution approaches is that it may yield better solutions than commonly used PDRs but with much shorter computation times and less formulation effort than optimal solvers [7]. Yet, DRL for scheduling problems is only in its infancy. On the one hand, the field still neglects some problem conditions inherent to real-world problems [8], and on the other hand it lags behind in the application of promising DRL paradigms such as curriculum learning (CL).

CL is a recent but very active research field in DRL and is built on the premise that, as with human learning, curricula play a critical role in effective learning behaviors in DRL. More precisely, CL is concerned with generating and learning from suitable experience sequences for the DRL agent. These sequences form the curriculum, which typically progressively varies the task difficulty leading up a final goal. The transfer of CL to the JSSP domain, has only recently been attempted [3]. Such existing methods design curriculums which vary between different problem sizes, i.e. numbers of jobs and machines per problem instance. While applicable to toy-box scenarios, the number of machines is often constant in real-world scenarios and corresponding usable training data. More granular CL within one fixed problem size, however, has not been studied yet. The missing component to accomplish CL in this granularity is a common definition of a degree of difficulty of problem instances within the same problem size.

In this work, we present such a definition and propose a new CL strategy for solving the JSSP with DRL. Comparing the learning behavior with and without CL, we empirically show the superiority of our approach with respect to the achieved average makespan. Our main contributions are summarized as follows:

- The introduction of a measure for the relative difficulty of a problem instance in JSSPs of the same problem size.
- A curriculum learning strategy for JSSPs suitable to steer the learning behavior of DRL agents and to receive shorter average makespans (compare Figure 1). The observed behavior shows that starting training on the most difficult instances decreases the resulting makespans by 3%.

The remainder of this paper is structured as follows: In section 2 we summarize latest achievements in DRL-based JSSP solutions and introduce CL in this context. Section 3 details our solution method and experimental setup, followed by the presentation of the results and insights in section 4 and their discussions in section 5. Finally, section 6 provides a conclusion and outlook to future work.

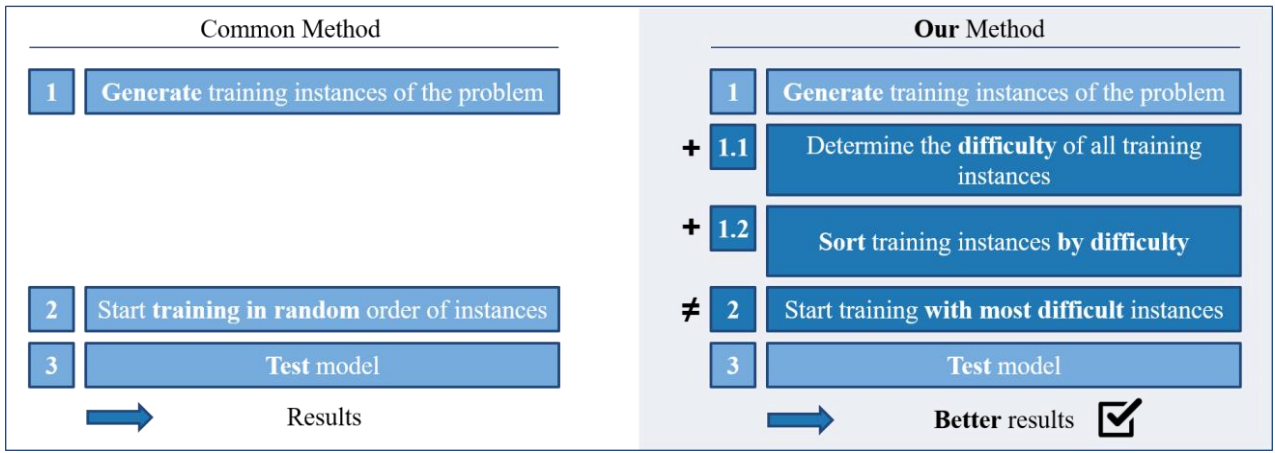

Figure 1: Comparison of the common method with our proposed method. Extension through calculations on the training instances and difference in training procedure.

## 2. Related Work

### 2.1 Deep Reinforcement Learning for Job Shop Scheduling Problems

Literature on DRL based JSSP solutions is rapidly increasing in volume and can roughly be divided into two classes: ground research and applied research. Ground research is generally concerned with new architectures [10,11,5], learning design decisions [3,4] and their comparison to existing solution methods, such as priority dispatching rules, meta heuristics and optimal solvers [12]. Here, we find continuously better performance of DRL on standard JSSP problems and benchmark datasets, matching and outperforming PDRs in recent years.

Applied research often considers an additional dimension in the problem formulation inspired by real-world use-cases, such as stochasticity [13,14], machine flexibility [15–17], dynamic job releases [18], machine failures [19] or multi-objective optimization criteria [20,21]. These studies show the general feasibility of DRL to learn, but are typically not very competitive with expert systems. Our contribution lies closer to the first class, as it methodologically extends an existing approach by means of a new learning paradigm for CL in job shop scheduling.

### 2.2 Curriculum Learning in Deep Reinforcement Learning based Job Shop Scheduling

According to Narvekar et al. [9], curriculum learning consists of three key elements: *task generation*, which deals with the division of the overall goal into easier sub-goals and the generation of suitable training experience; *sequencing*, dealing with the order in which to present the training experience; and *transfer learning*, comprising methods to tackle forgetting skills acquired from past experience when confronted with new experience.

CL for DRL-based JSSP is not much investigated in the current state of research. In a wider sense, CL is used in several approaches to DRL-based job shop scheduling by applying variations of experience replay [22–25,11,18], in which the gathered experience is rearranged and sampled aiming to make learning more stable. In that way, it is loosely related to the *sequencing* element of CL. However, experience replay works with the experience once it is already gathered, skipping the *task generation* element of CL. *Task generation* is less studied and a remaining challenge for solving combinatorial optimization problems with DRL [26]. In our work, we propose an own metric for the importance of experience based on the performance of priority dispatching rules, which serves as a discriminating factor for easy and hard tasks.

Iklassov et al. [3] explicitly propose CL in the JSSP domain. They define the easiness of sub-goals of JSSPs through problem sizes, as most common in combinatorial problems because of the solution space scales with the problem size [27]. By this definition, a problem instance with more jobs and machines is harder than one with less jobs and machines. Making use of a problem size agnostic neural network architecture, the authors introduce an automatic sequencing algorithm which favors collecting experiences from the currently hardest problem size. Their results indicate that models trained with CL drastically outperform those trained without CL. Our approach differs from that of Iklassov et al. [3] in that we apply CL for problem instances of the same size. Hence, we are closing an important gap that enables applying CL to those manufacturing scenarios in which the number of machines remains the same.

### 3. Methods

In the following, we first summarize the work by Zhang et al. [5], which serves as our methodological testbed and baseline, followed by the details of our CL extension and experimental setup.

### 3.1 Deep Reinforcement Learning Approach

Our approach extends the method and framework presented in Zhang et al. [5], which shows competitive results on recognized benchmark datasets of the JSSP with makespan optimization. Specifically, our method adapts the interaction logic of the DRL agent with the simulation (action-space and environment step), the action evaluation signal (reward), the input formulation (observation-space), and the network architecture.

The studied DRL agent iteratively plans tasks of a JSSP by choosing from the list of still unfinished jobs in each iteration step. The corresponding next task of this job is scheduled to start at the earliest still possible time by a mechanism called *left-shifting*: left shifting means that if the current plan, consisting of all scheduled tasks up to that point, can be optimized by switching the position of the chosen task with the previous one on the used machine, this switch is executed by the simulation. The corresponding reward signal consists of the difference of the makespan of the already scheduled tasks before and after the last step, such that the cumulative reward received throughout the planning process equals the negative makespan of the final plan. The scheduling decision is based on a size-agnostic embedding: For each task, the embedding contains the information whether it is done and what its current lower bound of the makespan is. Each task represents a node in a graph neural network in which the corresponding information is propagated from node to node and finally aggregated by summation.

In the original paper, the 40.000 training instances per agent were generated on the fly by randomly sorting processing orders on machines and drawing processing times randomly from a normal distribution. Our central extension is a different sampling procedure as part of the CL approach.

### 3.2 Curricular Training Procedure

**Task Generation (definition of instance difficulty):** In order to carry out curriculum learning, a feature to divide problem instances into subtasks that vary in difficulty is essential. Since instances of the same problem size by definition share the same computational complexity, we resort to a feature defined by how well we are already able to solve instances through an established set of rules, i.e. PDRs. We call this discriminative feature *difficulty to solve* (DTS). DTS is defined as the makespan, which the most competitive PDR achieves on any given problem instance. Accordingly, we speak of those instances on which a shorter than average makespan is realized through the best PDR as *easy* tasks and those on which a longer makespan is realized as *hard* tasks. Applied to our use case, we proceed as follows (cf. Figure 1, step 1 and 1.1 of our method): As in Zhang et al. [5], we generate 40.000 random 6x6 JSSP training instances from normal distributions with respect to machine orders and processing times. After solving the training data with six commonly used priority dispatching rules jointly with the left-shifting procedure used in Zhang et al. [5], we find that the *most tasks remaining* (MTR) prioritization rule performs best with an average of a 16% larger makespan compared to the optimal makespan (optimality gap). The results of all considered PDRs are shown in the appendix (Table A1). MTR only performed marginally better than the *least remaining processing time* (LRPT) prioritization, but much better than the most often used *shortest processing time* (SPT). Optimal solutions were generated using the CP-SAT solver by OR-Tools [25]. Figure 2a) depicts the distribution of achieved makespans through MTR, which is our used DTS metric.

**Sequencing:** The creation of training sequences is the next step. Often the difficulty is gradually increased over training in CL, following the intuition that an agent learns a basic strategy first and refines it later to match more difficult scenarios. To cover this sequence, but also others, we sort the training instances by DTS, as depicted in Figure 2b), then split it into the easy and hard halves and keep the original, or normal, order ($e\_n$, $h\_n$) or reverse it ($e\_r$, $h\_r$). For example, $e\_n$ (red line in Figure 2b)) consists of half of the training data in normal order, i.e. starting from the lowest DTS around 300 and ending at the mean DTS of about 580. The four portions make up our ordered training *curriculum elements* (CE). One entire training curriculum consists of two concatenated CEs, e.g. [$e\_n$, $e\_n$] or [$e\_n$, $h\_r$], resulting in the 16 possible curricula, schematically depicted in Figure 3.

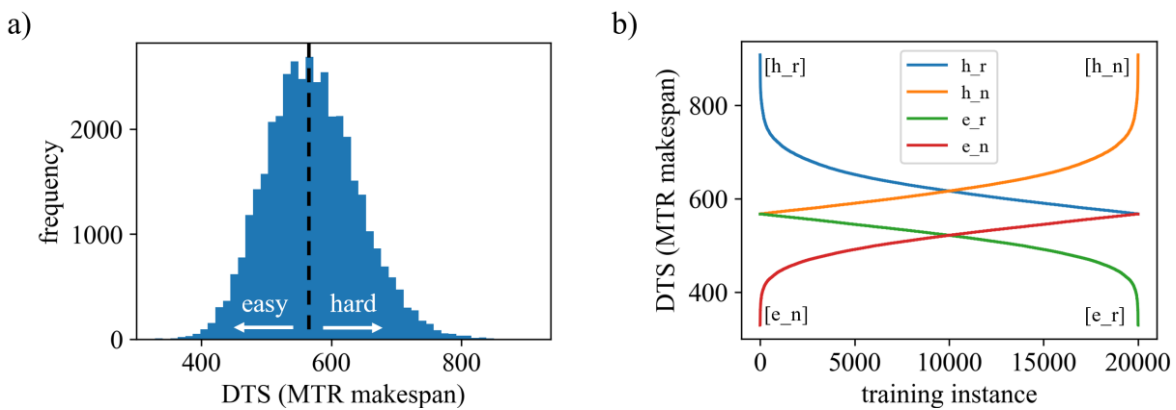

Figure 2 Training data consisting of 40.000 unique instances. a) Histogram of training instances by their DTS (makespan through the MTR dispatching rule); b) Elements of the curriculum: portions of training data sorted by DTS. (e_n = easy, normal order; e_r = easy, reversed order; h_n = hard, normal order; h_r = hard, reversed order)

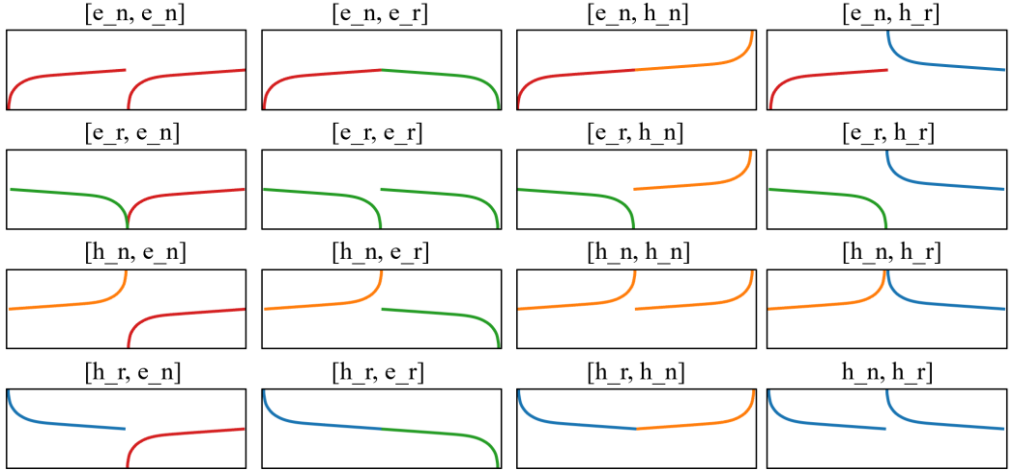

Figure 3: Schematic representation of all 16 possible curricula. As in Figure 2b), halves consist of CEs containing datapoints ordered by DTS (y-axis) along the training procedure (x-axis).

### 3.3 Experimental Setup

The experiments are designed such that differences in the agent behavior and performance may only be attributed to the training curricula. To this end, a separate RL-agent is trained for each of the 16 curricula until all training instances within the curriculum have been shown once. As baseline, we also train three RL-agents on unordered training data, where the problem instances were randomly shuffled and the agents are randomly initialized with varying random seeds. Training hyperparameters are fixed in accordance with Zhang et al. [5] for all experiments. All agents are tested on a fixed test dataset containing 1000 problem instances each time 2000 training instances have been shown. For more statistically significant results, we sampled three different training datasets with varying random seeds as described in section 3.2. The above experiments are carried out separately on all three datasets.

### 4. Experiment Results

Figure 4 shows the results of agents tested on the validation instances over the course of training. Agents trained on the same CE in the first half of the training period, e.g. on $e\_n$ in *[$e\_n$, $e\_n$]*, *[$e\_n$, $e\_r$]*, *[$e\_n$, $h\_n$]*, *[$e\_n$, $h\_r$]*, are averaged across the three datasets and depicted as solid lines. Generally, one can observe a rapid decline to a first dip of the optimality gap from the first validation point after 2.000 training instances to 6.000 training instances, followed by an increase in the optimality gap and a gradual subsequent convergence to final values towards the end of the training. Interestingly, more than 70% of the agents reach their global minimum in the first dip. This indicates that the agents develop the most successful strategy in the very beginning of training and never return to it, but converge towards a higher (worse) optimality gap instead. Moreover, the best agents 10% of all agents reached their minimum in the first dip.

A closer look at the first dip (cf. the zoom-in on the right in Figure 4) reveals that the lowest point is directly related to the easiness of the first data shown to the agent, hence the training curriculum. More precisely, the lowest point corresponds to agents trained with the $h\_r$ CE (blue line), meaning that they have been trained on the hardest training data. Inversely, agents trained on $e\_n$ (red line) remain highest among all points at the first dip. Noteworthily, all agents trained on a CE perform better than those trained without a curriculum (black line) on average. This means that it is advisable to use a curriculum, specifically the *h_r* curriculum in the beginning of the training, to achieve the best results. In our case, we achieve 7% better results (1.1%p) in the first dip. Overall, we achieve 3% better results (0.5%p).

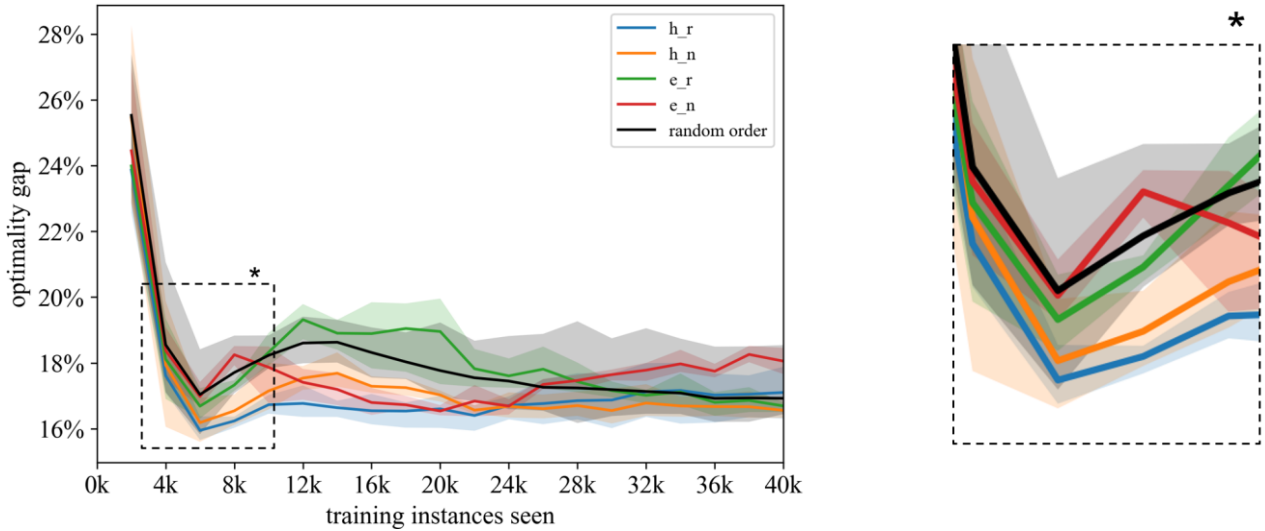

Figure 4: Performance of the agents on the test instances over training progress. Lines indicate the mean across three random seeds for the training instance generation. Shaded areas indicate the minimum and maximum values across three runs. Colored lines represent agents trained on curricula, where the first curriculum element is indicated in the legend. The second curriculum element is h_r for all depicted agents. The black line represents agents trained on randomly ordered training instances.

Next, we analyze the training behavior of the agents regarding the second half of training, where the second CEs are presented. Note that in some cases, such as *[e_n, e_n]*, the agent sees only one half of all instances. We study two main questions: Firstly, does the second CE have a consistent impact on the final result? Secondly, does the curriculum have a reproducible impact upon introduction in the middle of the training (difference between test after 20,000 and 22,000 training instances)? The latter may help to steer the agent away from a local minimum. Figure 5 shows the learning curves of agents trained on different curriculums composed from the same training dataset. In each plot, learning curves of four agents are displayed. The plots overlap in the first half of the training because of being trained on the same first CE, but diverge in the middle upon introduction of the second CE. Across all plots we were not able to find significant correlations between second curriculum elements and the learning curve with respect to optimal performance.

This answers the first question: the curriculum element in the second half does not have a consistent impact on the final overall result. However, we observed trends regarding the local behavior in the beginning of the second half of training. Similar to the behavior in the first half, *h_r* generally invokes the largest drop in optimality gap compared to the other CEs in three out of four cases. In some cases, this goes so far that while *h_r* invokes a drop in the optimality gap, *e_n* invokes a rise in the optimality gap.

Figure 6 visualizes the statistics of the immediate local impact of the CEs in the second half of training. Figure 6 a) shows the relative statistical impact of the CEs compared to the CEs by rank. The first bar indicates that in nine out of twelve cases, *h_r* invoked the largest immediate drop in optimality gap. Analogously, the last bar indicates that in five cases, *e_n* achieves the lowest performance. Generally, we find that *h_r* ranks highest and *e_n* lowest, whereas *h_n* and *e_r* rank in between. Similarly, in Figure 6 b), we can look at the absolute impact and count the number of times a CE caused an immediate jump towards better or worse optimality gap. Evidently *h_r* and *h_n* rather cause jumps towards better optimality gaps, whereas *e_r* is neutral and *e_n* causes jumps towards worse optimality gaps more often than not.

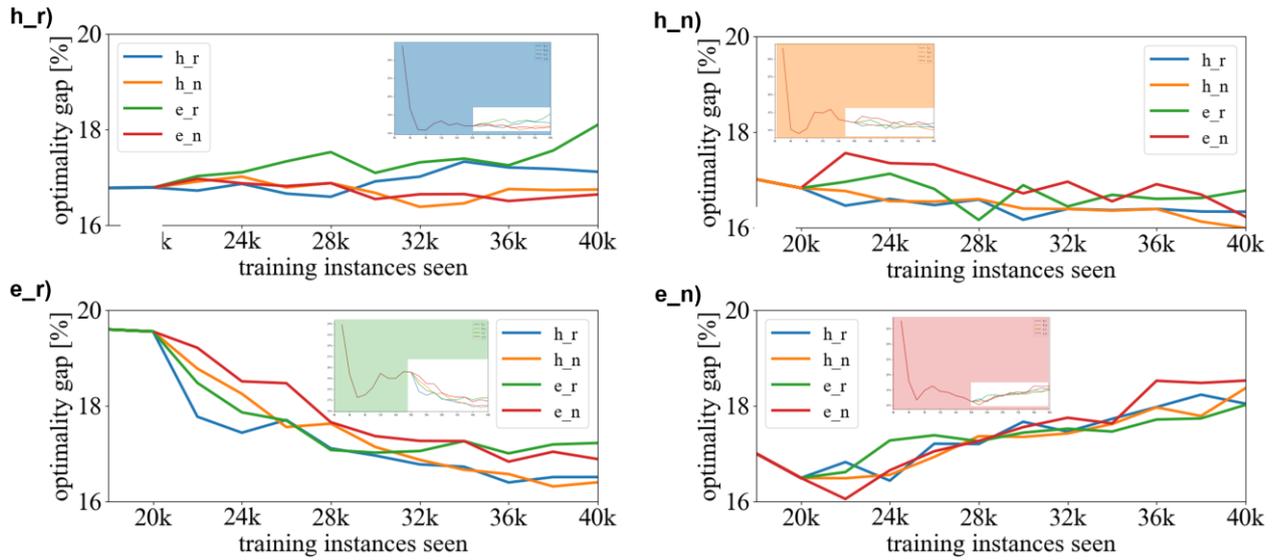

Figure 5: Zoom-ins on the second half of trainings. In each plot, training was performed on the same curriculum element in the first half (top left on [h_r], top right on [h_n], bottom left on [e_r] and bottom right on [e_n])

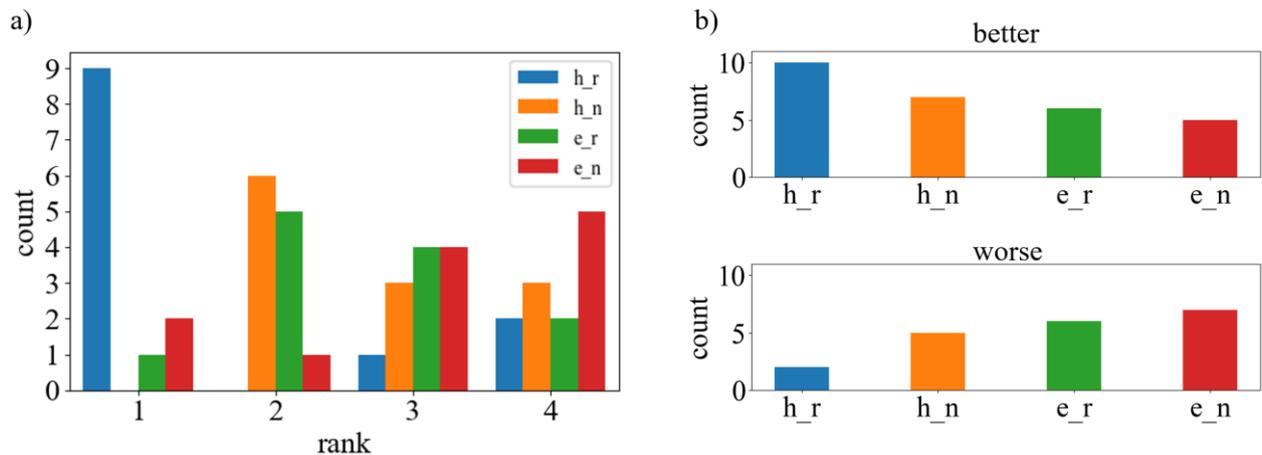

Figure 6: Statistical analysis of the local impact of training on first 2000 instances of each CE in the second half of training. a) Relative impact compared to other CEs. b) and c) Absolute immediate impact: count, how often each respective CE caused a drop (better) or rise (worse) in the optimality gap.

## 5. Discussion of Results

The presented results suggest that the learning behavior of the DRL agent can be positively influenced through the CEs defined in this study. As a practical consequence, we achieve better global results after a comparatively short training period. We therefore propose using CL according to our methodology, which is easily implemented and integrated into existing solution approaches. On a more fundamental level, the results suggest that the proposed DTS metric is useful to evaluate the easiness of a JSSP problem instance, a novum in this particular domain. During the experiments, we further observed the global minimum in the dip (cf. Figure 4) during training. Though useful in this particular case, in RL we much rather observe smooth, almost monotonically decreasing learning curves. An investigation for the reason behind the learning curve may be subject of future work.

Another noteworthy observation is that learning on the hardest problems first achieves the best outcomes. CL methods otherwise typically start from easier sub-problems and transfer this knowledge into the actual final problem. Our initial explanation attempt, is that the harder problems introduce a stronger negative reward signal through the larger makespan (note that our definition of DTS is related to the achieved makespan), pushing the agent more towards a certain initial strategy. Another intuitive hypothesis is that a strategy working well on harder problems, which inhibit a larger makespan, very effectively decreases the large optimality gap of these problems and leads to the strong results. To test whether a strategy that minimizes the particularly large optimality gaps may be incentivized through a curriculum, one may try using the optimality gap instead of the makespan achieved by MTR as DTS metric in the future. Note, however, that this requires solving every training instance optimally, which is much slower than our MTR-based approach especially when applying the method to larger problem instances.

## 6. Conclusion and Outlook

CL is a promising DRL paradigm, yet not well studied in the context of JSSP solutions. In this study we investigated the impact of a learning curriculum within a fixed problem size of the JSSP. We found that ordering training instances by how well an established priority dispatching rule, MTR, performs on these instances provides meaningful metric for forming curricula that allow us to improve the learning behavior of DRL agents and to increase the scheduling performance. By starting the training with instances sorted from worst to best performances of MTR, our approach consistently outperforms agents trained on randomly ordered training data.

Motivated by the presented results, in our future work we will investigate other metrics for the difficulty of problem instances of the same problem size. These may stem from priority dispatching rules that are combined for better performance or well suited for certain modifications of the JSSP. This is especially necessary for the successful transfer the methodology to other scheduling problems which include more challenging optimization objectives and additional constraints.

## Acknowledgements

This research work was undertaken within the research project AlphaMES funded by the German Federal Ministry for Economic Affairs and Climate Action (BMWK).

## Appendix

Table A1: Average optimality gap of priority dispatching rules on the training data.
SPT=Shortest Processing Time First (job-wise); LPT=Longest Processing Time First (job-wise); MTR=Most Tasks Remaining (job-wise); LRPT=Least Remaining Processing (job-wise); MPTLOM=Most Processing Time Left On Machine (machine-wise); RANDOM: Random Prioritization of Jobs

| PDR | SPT | LPT | **MTR** | LRPT | LOUM | MPTLOM | RANDOM |
|---|---|---|---|---|---|---|---|
| opt. gap | 0.40 | 0.32 | **0.16** | 0.16 | 0.41 | 0.35 | 0.29 |

**Biography**

**Constantin Waubert de Puiseau** (*1994) holds a Master's Degree from RWTH Aachen University in Mechanical Engineering. Since 2019, he works at the chair of Digital Transformation Technologies and Management at the University of Wuppertal. His research focuses on solving real-world planning and scheduling problems with reinforcement learning.

**Hasan Tercan** (*1988) holds a Master's Degree from TU Darmstadt in Computer Science. Since 2018, he is a scientific researcher at the chair of Technologies and Management of Digital Transformation and leader of the research group Industrial Deep Learning. In his research, he investigates the application of machine learning methods in industrial processes.

**Tobias Meisen** (*1981) is a Professor of Digital Transformation Technologies and Management at the University of Wuppertal since 2018. He is also the Institute Director of the In-Institute for Systems Research in Information, Communication and Media Technology, the vice-chair of the Interdisciplinary Centre for Data Analytics and Machine Learning and a co-founder of Hotsprings GmbH that is now part of Umlaut SE.